\documentclass[10pt,twocolumn,letterpaper]{article}

\usepackage[pagenumbers]{cvpr} 

\usepackage{graphicx}
\usepackage{amsmath,amsthm}
\usepackage{amssymb}
\usepackage{booktabs}
\usepackage{xcolor,soul,verbatim}
\newcommand{\cbox}[2]{\colorbox{#1}{$\displaystyle #2$}}

\usepackage[pagebackref,breaklinks,colorlinks]{hyperref}

\newtheorem{thm}{Theorem}[section]
\newtheorem{dfn}[thm]{Definition}
\newtheorem{pro}[thm]{Problem}

\newtheorem{cor}[thm]{Corollary}
\newtheorem{lem}[thm]{Lemma}
\newtheorem{exa}[thm]{Example}

\newcommand{\R}{\mathbb R}

\newcommand{\la}{\lambda}

\newcommand{\si}{\sigma}

\newcommand{\ep}{\varepsilon}

\newcommand{\Or}{\mathrm{O}}
\newcommand{\HKS}{\mathrm{HKS}}

\newcommand{\ADD}{\mathrm{ADD}}
\newcommand{\ASD}{\mathrm{ASD}}
\newcommand{\SDM}{\mathrm{SDM}}
\newcommand{\CDM}{\mathrm{CDM}}
\newcommand{\SDV}{\mathrm{SDV}}
\newcommand{\CID}{\mathrm{CID}}

\newcommand{\LAC}{\mathrm{LAC}}
\newcommand{\PDD}{\mathrm{PDD}}

\newcommand{\SDD}{\mathrm{SDD}}

\newcommand{\RDD}{\mathrm{RDD}}

\newcommand{\ORD}{\mathrm{ORD}}

\newcommand{\OCD}{\mathrm{OCD}}

\newcommand{\SCD}{\mathrm{SCD}}
\newcommand{\mSCD}{\overline{\SCD}}

\newcommand{\ACD}{\mathrm{ACD}}
\newcommand{\ACV}{\mathrm{ACV}}
\newcommand{\AMD}{\mathrm{AMD}}

\newcommand{\EMD}{\mathrm{EMD}}

\newcommand{\sign}{\mathrm{sign}}

\newcommand{\lra}{\leftrightarrow}
\newcommand{\Lra}{\Leftrightarrow}
\newcommand{\es}{\emptyset}

\newcommand{\pd}[2]{\frac{\partial #1}{\partial #2}}

\newcommand{\vect}[2]{ \left( \begin{array}{c} 
 #1 \\ #2 \end{array} \right)}

\newcommand{\mat}[4]{ \left( \begin{array}{cc} 
 #1 & #2 \\ #3 & #4 \end{array} \right)}
\newcommand{\matv}[4]{ \left( \begin{array}{cc} 
 #1 & #3 \\ #2 & #4 \end{array} \right)}

\usepackage[capitalize]{cleveref}
\crefname{section}{Sec.}{Secs.}
\Crefname{section}{Section}{Sections}
\Crefname{table}{Table}{Tables}
\crefname{table}{Tab.}{Tabs.}

\def\cvprPaperID{11825} 
\def\confName{CVPR}
\def\confYear{2023}

\begin{document}

\title{Recognizing Rigid Patterns of Unlabeled Point Clouds by Complete and Continuous Isometry Invariants with no False Negatives and no False Positives}

\author{Daniel Widdowson\\
Computer Science department\\
University of Liverpool, UK\\
{\tt\small d.e.widdowson@liverpool.ac.uk}
\and
Vitaliy Kurlin\\
Computer Science department\\
University of Liverpool, UK\\
{\tt\small vitaliy.kurlin@gmail.com}
}
\maketitle

\begin{abstract}
Rigid structures such as cars or any other solid objects are often represented by finite clouds of unlabeled points.
The most natural equivalence on these point clouds is rigid motion or isometry maintaining all inter-point distances.
\smallskip

Rigid patterns of point clouds can be reliably compared only by complete isometry invariants that can also be called equivariant descriptors without false negatives (isometric clouds having different descriptions) and without false positives (non-isometric clouds with the same description).
\smallskip

Noise and motion in data motivate a search for invariants that are continuous under perturbations of points in a suitable metric.
We propose the first continuous and complete invariant of unlabeled clouds in any Euclidean space.
For a fixed dimension, the new metric for this invariant is computable in a polynomial time in the number of points.
\end{abstract}

\section{Strong motivations for complete invariants}
\label{sec:intro}

In Computer Vision, real objects such as cars and solid obstacles are considered rigid and often represented by a finite set $C\subset\R^n$ (called a \emph{cloud}) of $m$ \emph{unlabeled} (or unordered) points, usually in low dimensions $n=2,3,4$. 
\smallskip

The rigidity of many real objects motivates the most fundamental equivalence of \emph{rigid motion} \cite{wang2019deep}, a composition of translations and rotations in $\R^n$.
In a general metric space $M$, the most relevant equivalence is \emph{isometry}: any map $M\to M$ maintaining all inter-point distances in $M$.
\smallskip

Any isometry in $\R^n$ is a composition of a mirror reflection with some rigid motion.
Any orientation-preserving isometry can be realized as a continuous rigid motion.
\smallskip

There is no sense in distinguishing rigid objects that are related by isometry or having the same shape.
Formally, the \emph{shape} of a cloud $C$ is its isometry class \cite{pomerleau2015review} defined as a collection of all infinitely many clouds isometric to $C$. 
\smallskip

The only reliable tool for distinguishing clouds up to isometry is an \emph{invariant} defined as a function or property preserved by any isometry.
Since any isometry is bijective, the number of points is an isometry invariant, but the coordinates of points are not invariants even under translation. 
This simple invariant is \emph{incomplete} (non-injective) because non-isometric clouds can have different numbers of points.
\smallskip

Any invariant $I$ maps all isometric clouds to the same value.
There are no isometric clouds $C\cong C'$ with $I(C)\neq I(C')$, meaning that $I$ has \emph{no false negatives}.
Isometry invariants are also called \emph{equivariant descriptors} \cite{schmidt2012learning}.
\smallskip

A \emph{complete} invariant $I$ should distinguish all non-isometric clouds, so if $C\not\cong C'$ then $I(C)\neq I(C')$.
Equivalently, if $I(C)=I(C')$ then $C\cong C'$, so $I$ has \emph{no false positives}.
Then $I$ can be considered as a DNA-style code or genome that identifies any cloud uniquely up to isometry. 
\smallskip

Since real data is always noisy and motions of rigid objects are important to track, a useful complete invariant must be also continuous under the movement of points. 
\smallskip

\begin{figure}[h!]
\centering
\includegraphics[width=\linewidth]{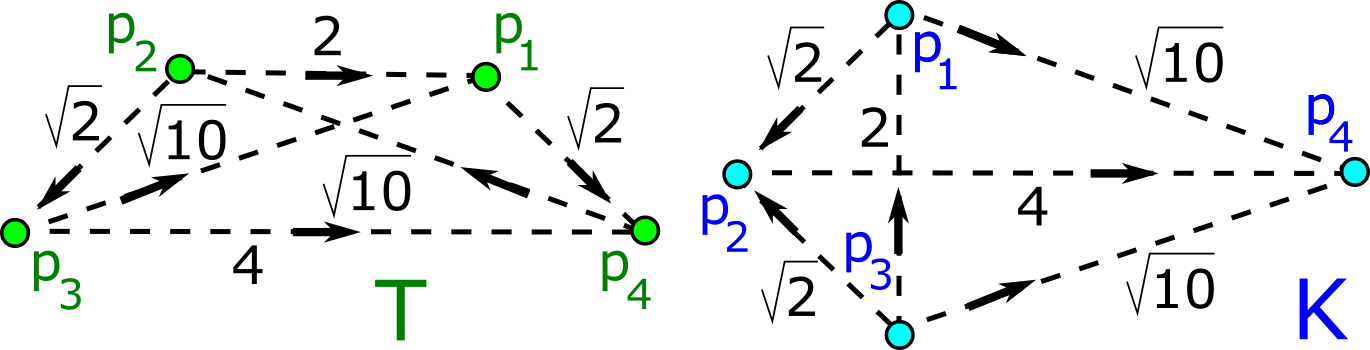}
\caption{
\textbf{Left}: the cloud $T=\{(1,1),(-1,1),(-2,0),(2,0)\}$. 
\textbf{Right}: the kite $K=\{(0,1),(-1,0),(0,-1),(3,0)\}$. 
$T$ and $K$ have the same 6 pairwise distances $\sqrt{2},\sqrt{2},2,\sqrt{10},\sqrt{10},4$.
}
\label{fig:4-point_clouds}
\end{figure}

A complete and continuous invariant for $m=3$ points consists of three pairwise distances (sides of a triangle) and is known in school as the SSS theorem \cite{weisstein2003triangle}.
But all pairwise distances are incomplete for $m\geq 4$ \cite{boutin2004reconstructing}, see Fig.~\ref{fig:4-point_clouds}.

\begin{pro}[complete isometry invariants with computable continuous metrics]
\label{pro:isometry}
For any cloud of $m$ unlabeled points in $\R^n$,
find an invariant $I$ satisfying the properties
\medskip

\noindent
\textbf{(a)}
\emph{completeness} : 
$C,C'$ are isometric $\Lra$ $I(C)=I(C')$; 
\medskip

\noindent
\textbf{(b)}
\emph{Lipschitz continuity} :
if any point of $C$ is perturbed within its $\ep$-neighborhood then
 $I(C)$ changes by at most $\la\ep$ for a constant $\la$ and a metric $d$ 
 satisfying these axioms:
\smallskip

\noindent 
1) $d(I(C),I(C'))=0$ if and only if $C\cong C'$ are isometric,
\smallskip

\noindent 
2) \emph{symmetry} : $d(I(C),I(C'))=d(I(C'),I(C))$,
\smallskip

\noindent 
3) $d(I(C),I(C'))+d(I(C'),I(C''))\geq d(I(C),I(C''))$;
\medskip

\noindent
\textbf{(c)}
\emph{computability} : $I$ and $d$ are computed in a polynomial time in the number $m$ of points for a fixed dimension $n$.
\end{pro}

Condition~(\ref{pro:isometry}b) asking for a continuous metric is stronger than the completeness in (\ref{pro:isometry}a).
Detecting an isometry $C\cong C'$ gives a discontinuous metric, say $d=1$ for all non-isometric clouds $C\not\cong C'$ even if $C,C'$ are nearly identical.
Any metric $d$ satisfying the first axiom in (\ref{pro:isometry}b) detects an isometry $C\cong C'$ by checking if $d=0$.
\smallskip

Theorem~\ref{thm:SCD} will solve Problem~\ref{pro:isometry} for any $m$ in $\R^n$.
Continuous invariants in Theorem~\ref{thm:SDD} are conjectured to be complete (no known counter-examples) in any metric space.
The first author implemented all algorithms, the second author wrote all theory, proofs, examples in \cite{kurlin2023simplexwise,kurlin2023strength}.

\section{Past work on cloud recognition/classification}
\label{sec:past}

\noindent  
\textbf{Labeled clouds} $C\subset\R^n$ are easy for isometry classification because the matrix of distances $d_{ij}$ between indexed points $p_i,p_j$ allows us to reconstruct $C$ by using the known distances to the previously constructed points \cite[Theorem~9]{grinberg2019n}.
For any clouds of the same number $m$ of labeled points, the difference between $m\times m$ matrices of distances (or Gram matrices of $p_i\cdot p_j$) can be converted into a continuous metric by taking a matrix norm.
If the given points are unlabeled, comparing $m\times m$ matrices requires $m!$ permutations, which makes this approach impractical.
\smallskip

\noindent
\textbf{Multidimensional scaling} (MDS).
For a given $m\times m$ distance matrix of any $m$-point cloud $A$, MDS \cite{schoenberg1935remarks} finds an embedding $A\subset\R^k$ (if it exists) preserving all distances of $M$ for a dimension $k\leq m$.
A final embedding $A\subset\R^k$ uses 
eigenvectors whose ambiguity up to signs gives an exponential comparison time that can be close to $O(2^m)$.
\smallskip

\noindent  
\textbf{Isometry detection} refers to a simpler version of Problem~\ref{pro:isometry} to algorithmically detect a potential isometry between given clouds of $m$ points in $\R^n$. 
The best algorithm by Brass and Knauer \cite{brass2000testing} takes $O(m^{\lceil n/3\rceil}\log m)$ time, so $O(m\log m)$ in $\R^3$ \cite{brass2004testing}.
These algorithms output a binary answer (yes/no) without quantifying similarity between non-isometric clouds by a continuous metric.
\smallskip

\noindent
\textbf{The Hausdorff distance} \cite{hausdorff1919dimension} can be defined for any subsets $A,B$ in an ambient metric space as  $d_H(A,B)=\max\{\vec d_H(A,B), \vec d_H(B,A) \}$, where the directed Hausdorff distance is 
$\vec d_H(A,B)=\sup\limits_{p\in A}\inf\limits_{q\in B}|p-q|$.
To take into account isometries, one can minimize the Hausdorff distance over all isometries \cite{huttenlocher1993comparing,chew1992improvements,chew1999geometric}.
For $n=2$, the Hausdorff distance minimized over isometries in $\R^2$ for sets of at most $m$ point needs $O(m^5\log m)$ time \cite{chew1997geometric}. 
For a given $\ep>0$ and $n>2$, the related problem to decide if $d_H\leq\ep$ up to translations has the time complexity $O(m^{\lceil(n+1)/2\rceil})$ \cite[Chapter~4, Corollary~6]{wenk2003shape}. 
For general isometry, only approximate algorithms tackled minimizations for infinitely many rotations initially in $\R^3$ \cite{goodrich1999approximate} and in $\R^n$ \cite[Lemma~5.5]{anosova2022algorithms}. 
\smallskip

\noindent
\textbf{The Gromov-Wasserstein distances} can be defined for metric-measure spaces, not necessarily sitting in a common ambient space.
The simplest Gromov-Hausdorff (GH) distance cannot be approximated with any factor less than 3 in polynomial time unless P = NP \cite[Corollary~3.8]{schmiedl2017computational}.
Polynomial-time algorithms for GH were designed for ultrametric spaces  \cite{memoli2021gromov}.
However, GH spaces are challenging even for point clouds sets in $\R$, see \cite{majhi2019approximating} and \cite{zava2023gromov}.
\smallskip

\noindent
\textbf{The Heat Kernel Signature} ($\HKS$) 
is a complete isometry invariant of a manifold $M$ whose the Laplace-Beltrami operator has distinct eigenvalues by \cite[Theorem~1]{sun2009concise}.
If $M$ is sampled by points, $\HKS$ can be discretized and remains continuous \cite[section~4]{sun2009concise} but the completeness is unclear.
\smallskip

\noindent  
\textbf{Equivariant descriptors} can be experimentally optimized
\cite{nigam2022equivariant,simeonov2022neural} on big datasets of clouds that are split into pre-defined clusters.
Using more hidden parameters can improve accuracy on any finite dataset at a higher cost but will require more work for any new data.
Point cloud registration filters outliers \cite{shi2021robin}, samples rotations for Scale Invariant Feature Transform or uses a basis \cite{toews2013efficient,rister2017volumetric,spezialetti2019learning,zhu2022point}, which can be unstable under perturbations of a cloud.
The PCA-based complete invariant of unlabelled clouds \cite{kurlin2022computable} can discontinuously change when a basis degenerates to a lower dimensional subspace but inspired Complete Neural Networks \cite{hordan2023complete} though without the Lipschitz continuity.
\smallskip

\noindent  
\textbf{Geometric Deep Learning} produces descriptors that are equivariant by design \cite{bronstein2021geometric} and go beyond Euclidean space $\R^n$ \cite{bronstein2017geometric}, hence aiming to experimentally solve Problem~\ref{pro:isometry}.
Motivated by obstacles in \cite{dong2018boosting,akhtar2018threat,laidlaw2019functional,guo2019simple,colbrook2022difficulty}, Problem~\ref{pro:isometry} needs a justified solution without relying on finite data.
\smallskip

\noindent  
\textbf{Geometric Data Science} solves analogs of Problem~\ref{pro:isometry} for any real data objects considered up to practical equivalences instead of rigid motion on clouds \cite{smith2022families,elkin2020mergegram,elkin2021isometry}:
1-periodic discrete series \cite{anosova2022density,anosova2023density,kurlin2022computable},
2D lattices \cite{kurlin2022mathematics,bright2023geographic}, 3D lattices \cite{mosca2020voronoi,bright2021welcome,kurlin2022exactly,kurlin2022complete}, periodic point sets in $\R^3$ \cite{smith2022practical,edelsbrunner2021density} and in higher dimensions \cite{anosova2021introduction,anosova2021isometry,anosova2022algorithms}.
The applications of  to crystalline materials  \cite{ropers2022fast,balasingham2022compact,vriza2022molecular,zhu2022analogy}
led to the \emph{Crystal Isometry Principle} \cite{widdowson2022average,widdowson2021pointwise,widdowson2022resolving}
extending Mendeleev's table of elements to the \emph{Crystal Isometry Space} of all periodic crystals parametrised by complete invariants like a geographic map of a planet.
\smallskip

\noindent
\textbf{Local distributions of distances} in M\'emoli's seminal work \cite{memoli2011gromov,memoli2022distance} for metric-measure spaces, or shape distributions \cite{belongie2002shape, grigorescu2003distance, manay2006integral, pottmann2009integral}, are first-order versions of the new $\SDD$ below.

\section{Simplexwise Distance Distribution (SDD)}
\label{sec:SDD}

\begin{figure}[h!]
\centering
\includegraphics[width=\linewidth]{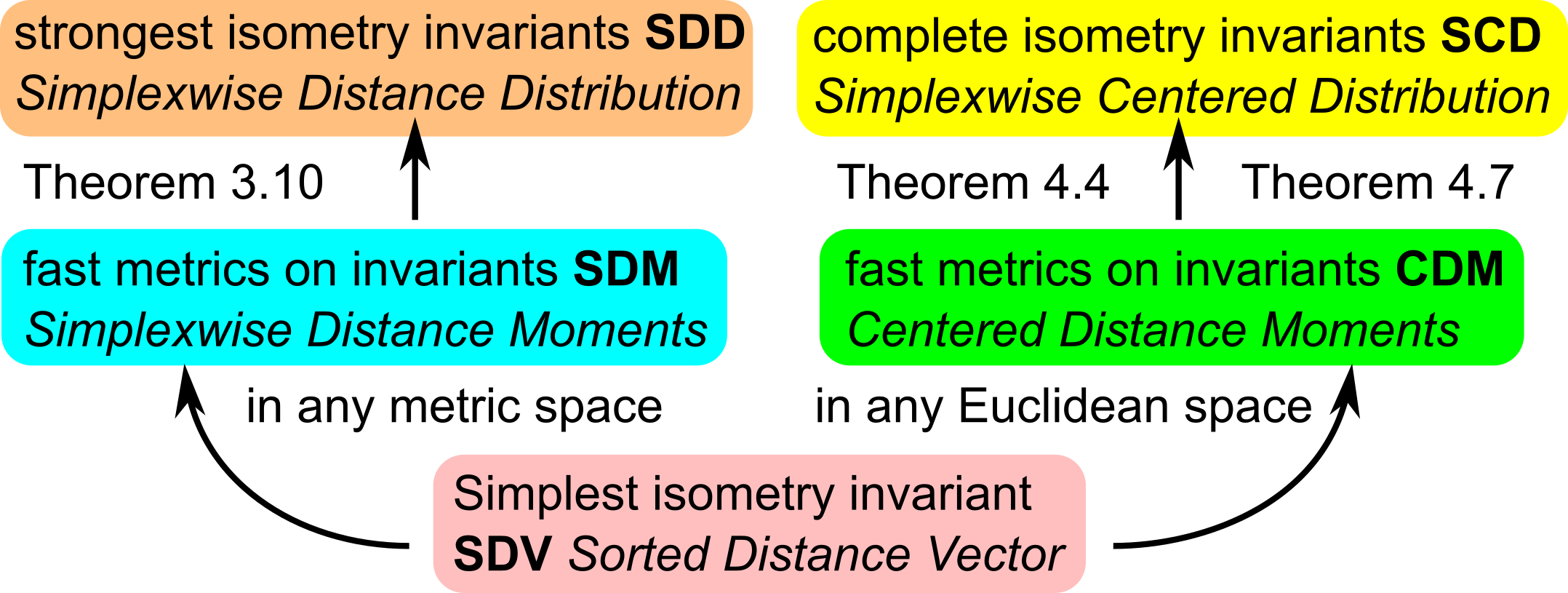}
\caption{Hierarchy of new invariants on top of the classical $\SDV$.}
\label{fig:invariants+diagram}
\end{figure}

We will refine Sorted Distance Vector in any metric space to get a complete invariant in $\R^n$ as shown in Fig.~\ref{fig:invariants+diagram}.
All proofs from sections~\ref{sec:SDD} and~\ref{sec:SCD} are in \cite{kurlin2023simplexwise,kurlin2023strength}, respectively.

\noindent
The \emph{lexicographic} order $u<v$ on vectors $u=(u_1,\dots,u_h)$ and $v=(v_1,\dots,v_h)$ in $\R^h$ means that if the first $i$ (possibly, $i=0$) coordinates of $u,v$ coincide then $u_{i+1}<v_{i+1}$.
Let $S_h$ denote the permutation group on indices $1,\dots,h$.

\begin{dfn}[$\RDD(C;A)$] 
\label{dfn:RDD}
Let $C$ be a cloud of $m$ unlabeled points in a space with a metric $d$.
Let $A=(p_1,\dots,p_h)\subset C$ be an ordered subset of $1\leq h<m$ points.
Let $D(A)$ be the \emph{triangular distance} matrix whose entry $D(A)_{i,j-1}$ is $d(p_i,p_j)$ for $1\leq i<j\leq h$, all other entries are filled by zeros.
Any permutation $\xi\in S_h$ acts on $D(A)$ by mapping $D(A)_{ij}$  to $D(A)_{kl}$, where $k\leq l$ is the pair of indices $\xi(i),\xi(j)-1$ written in increasing order.
\smallskip

For any other point $q\in C-A$, write distances from $q$ to $p_1,\dots,p_h$ as a column.
The $h\times (m-h)$-matrix $R(C;A)$ is formed by these $m-h$ lexicographically ordered columns.
The action of $\xi$ on $R(C;A)$ maps any $i$-th row to the $\xi(i)$-th row, after which all columns can be written again in the lexicographic order.
The \emph{Relative Distance Distribution} $\RDD(C;A)$ is the equivalence class of the pair $[D(A),R(C;A)]$ of matrices up to  permutations $\xi\in S_h$.
\end{dfn}

For a 1-point subset $A=\{p_1\}$ with $h=1$, the matrix $D(A)$ is empty and $R(C;A)$ is a single row of distances (in the increasing order) from $p_1$ to all other points $q\in C$.
For a 2-point subset $A=(p_1,p_2)$ with $h=2$, the matrix $D(A)$ is the single number $d(p_1,p_2)$ and $R(C;A)$ consists of two rows of distances from $p_1,p_2$ to all other points $q\in C$.

\begin{figure}[h!]
\centering
\includegraphics[width=\linewidth]{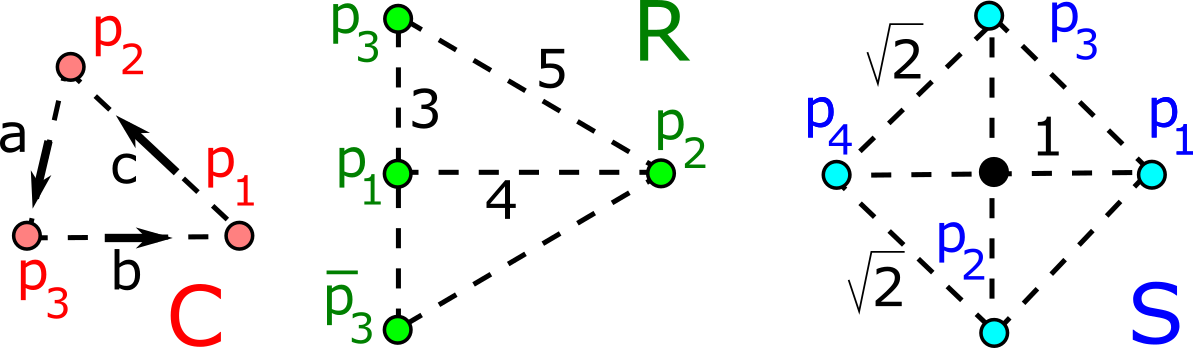}
\caption{\textbf{Left}: a cloud $C=\{p_1,p_2,p_3\}$ with distances $a\leq b\leq c$.
\textbf{Middle}: the triangular cloud $R=\{(0,0),(4,0),(0,3)\}$. 
\textbf{Right}: the square cloud $S=\{(1,0),(-1,0),(0,1),(-1,0)\}$. 
}
\label{fig:triangular_clouds}
\end{figure}

\begin{exa}[$\RDD$ for a 3-point cloud $C$]
\label{exa:RDD}
Let $C\subset\R^2$ consist of $p_1,p_2,p_3$ with inter-point distances $a\leq b\leq c$ ordered counter-clockwise as in Fig.~\ref{fig:triangular_clouds}~(left).  
Then
$$\RDD(C;p_1)=[\es;(b,c)], 
\RDD(C;\vect{p_2}{p_3})=[a;\vect{c}{b}],$$
$$\RDD(C;p_2)=[\es;(a,c)], 
\RDD(C;\vect{p_3}{p_1})=[b;\vect{a}{c}],$$
$$\RDD(C;p_3)=[\es;(a,b)], 
\RDD(C;\vect{p_1}{p_2})=[c;\vect{b}{a}].$$
We will always represent $\RDD$ for a specified order $A=(p_i,p_j)$ of points that are written as a column.
Swapping the points $p_1\lra p_2$ makes the last $\RDD$ above equivalent to another form:
$\RDD(C;\vect{p_2}{p_1})=[c;\vect{a}{b}]$.
\end{exa}

Though $\RDD(C;A)$ is defined up to a permutation $\xi$ of $h$ points in $A\subset C$, we later use only $h=n$, which makes comparisons of $\RDD$s practical in dimensions $n=2,3$.
Metrics on isometry classes of $C$ will be independent of $\xi$. 

\begin{dfn}[Simplexwise Distance Distribution $\SDD(C;h)$]
\label{dfn:SDD}
Let $C$ be a cloud of $m$ unlabeled points in a metric space.
For an integer $1\leq h<m$, the \emph{Simplexwise Distance Distribution} $\SDD(C;h)$ is the unordered set of $\RDD(C;A)$ for all unordered $h$-point subsets $A\subset C$.
\end{dfn}

For $h=1$ and any $m$-point cloud $C$, the distribution $\SDD(C;1)$ can be considered as a matrix of $m$ rows of ordered distances from every point $p\in C$ to all other $m-1$ points.
If we lexicographically order these $m$ rows and collapse any $l>1$ identical rows into a single one with the weight $l/m$, then we get the
Pointwise Distance Distribution $\PDD(C;m-1)$ introduced in \cite[Definition~3.1]{widdowson2022resolving}. 
\smallskip

The PDD was simplified to the easier-to-compare vector of Average Minimum Distances \cite{widdowson2022average}:
$\AMD_k(C)=\dfrac{1}{m}\sum\limits_{i=1}^m d_{ik}$, where $d_{ik}$ is the distance from a point $p_i\in C$ to its $k$-th nearest neighbor in $C$.
These neighbor-based invariants can be computed in a near-linear time in $m$ \cite{elkin2022new} and were pairwise compared for all all 660K+ periodic crystals in the world's largest database of real materials \cite{widdowson2022resolving}. 
Definition~\ref{dfn:SDM} similarly maps $\SDD$ to a smaller invariant.
\smallskip

Recall that the 1st moment of a set of numbers $a_1,\dots,a_k$ is the \emph{average} $\mu=\dfrac{1}{k}\sum\limits_{i=1}^k a_i$.
The 2nd moment is the \emph{standard deviation} $\si=\sqrt{\dfrac{1}{k}\sum\limits_{i=1}^k (a_i-\mu)^2}$.
For $l\geq 3$, the $l$-th \emph{standardized moment} \cite[section~2.7]{keeping1995introduction} is $\dfrac{1}{k}\sum\limits_{i=1}^k \left(\dfrac{a_i-\mu}{\si}\right)^l$.
 
\begin{dfn}[Simplexwise Distance Moments $\SDM$]
\label{dfn:SDM}
For any $m$-point cloud $C$ in a metric space,
 let $A\subset C$ be a subset of $h$ unordered points.
The \emph{Sorted Distance Vector} $\SDV(A)$ is the list of all $\frac{h(h-1)}{2}$ pairwise distances between points of $A$ written in increasing order. 
The vector $\vec R(C;A)\in\R^{m-h}$ is obtained from the $h\times(m-h)$ matrix $R(C;A)$ in Definition~\ref{dfn:RDD} by writing the vector of $m-h$ column averages in increasing order.
\smallskip

The pair $[\SDV(A);\vec R(C;A)]$ is the \emph{Average Distance Distribution} $\ADD(C;A)$ considered as a vector of length $\frac{h(h-3)}{2}+m$.
The unordered collection of $\ADD(C;A)$ for all $\binom{m}{h}$ unordered subsets $A\subset C$ is the Average Simplexwise Distribution $\ASD(C;h)$.
The \emph{Simplexwise Distance Moment} $\SDM(C;h,l)$ is the $l$-th (standardized for $l\geq 3$) moment of $\ASD(C;h)$ considered as a probability distribution of $\binom{m}{h}$ vectors, separately for each coordinate.
\end{dfn}

\begin{exa}[$\SDD$ and $\SDM$ for $T,K$]
\label{exa:SDD}
Fig.~\ref{fig:4-point_clouds} shows the non-isometric 4-point clouds $T,K$ with the same Sorted Distance Vector $\SDV=\{\sqrt{2},\sqrt{2},2,\sqrt{10},\sqrt{10},4\}$, see infinitely many examples in \cite{boutin2004reconstructing}.
The arrows on the edges of $T,K$ show orders of points in each pair of vertices for $\RDD$s.
Then $T,K$ are distinguished up to isometry by $\SDD(T;2)\neq \SDD(K;2)$ in Table~\ref{tab:SDD+TK}.
The 1st coordinate of $\SDM(C;2,1)\in\R^3$ is the average of 6 distances from $\SDV(T)=\SDV(K)$ but the other two coordinates (column averages from $R(C;A)$ matrices) differ.
\end{exa}

\begin{table}
  \centering
  \begin{tabular}{@{}l|l@{}}
    \toprule
   	$\RDD(T;A)$ in $\SDD(T;2)$ & $\RDD(K;A)$ in $\SDD(K;2)$ \\
    
    	$[\sqrt{2},\matv{2}{\cbox{yellow}{\sqrt{10}}}{\sqrt{10}}{4}]\times 2$ &
	$[\sqrt{2},\matv{2}{\cbox{yellow}{\sqrt{2}}}{\sqrt{10}}{4}]\times 2$ 
	\smallskip \\
	
$[2,\matv{\sqrt{2}}{\cbox{yellow}{\sqrt{10}}}{\sqrt{10}}{\cbox{yellow}{\sqrt{2}}}]$ & 
$[2,\matv{\sqrt{2}}{\cbox{yellow}{\sqrt{2}}}{\sqrt{10}}{\cbox{yellow}{\sqrt{10}}}]$ 
\smallskip \\
    
	$[\sqrt{10},\matv{\sqrt{2}}{\cbox{yellow}{2}}{\cbox{yellow}{4}}{\cbox{yellow}{\sqrt{2}}}]\times 2$ &
	$[\sqrt{10},\matv{\sqrt{2}}{\cbox{yellow}{4}}{\cbox{yellow}{2}}{\cbox{yellow}{\sqrt{10}}}]\times 2$ \smallskip \\
	
	    $[4,\matv{\sqrt{2}}{\sqrt{10}}{\cbox{yellow}{\sqrt{10}}}{\cbox{yellow}{\sqrt{2}}}]$ &
	$[4,\matv{\sqrt{2}}{\sqrt{10}}{\cbox{yellow}{\sqrt{2}}}{\cbox{yellow}{\sqrt{10}}}]$  \\

    \midrule
  	$\ADD(T;A)$ in $\ASD(T;2)$ & $\ADD(K;A)$ in $\ASD(K;2)$ \\
    $[\sqrt{2},(\frac{2+\cbox{yellow}{\sqrt{10}}}{2},\frac{4+\sqrt{10}}{2})]\times 2$ &
    $[\sqrt{2},(\frac{2+\cbox{yellow}{\sqrt{2}}}{2},\frac{4+\sqrt{10}}{2})]\times 2$ \\

    $[2,(\cbox{yellow}{\frac{\sqrt{2}+\sqrt{10}}{2},\frac{\sqrt{2}+\sqrt{10}}{2}})]$ &
    $[2,(\cbox{yellow}{\sqrt{2},\sqrt{10}})]$ \\

    $[\sqrt{10},(\frac{2+\cbox{yellow}{\sqrt{2}}}{2},\frac{4+\sqrt{2}}{2})]\times 2$ &
    $[\sqrt{10},(\frac{2+\cbox{yellow}{\sqrt{10}}}{2},\frac{4+\sqrt{2}}{2})]\times 2$ \\
    
       $[4,(\frac{\sqrt{2}+\sqrt{10}}{2},\frac{\sqrt{2}+\sqrt{10}}{2})]$ & $[4,(\frac{\sqrt{2}+\sqrt{10}}{2},\frac{\sqrt{2}+\sqrt{10}}{2})]$     \\

    \midrule
   	$\SDM_1=\dfrac{3+\sqrt{2}+\sqrt{10}}{3}$ & 
   	$\SDM_1=\dfrac{3+\sqrt{2}+\sqrt{10}}{3}$ \\

   	$\SDM_2=\dfrac{\cbox{yellow}{6+2\sqrt{2}+4\sqrt{10}}}{12}$ & 
   	$\SDM_2=\dfrac{\cbox{yellow}{8+5\sqrt{2}+3\sqrt{10}}}{12}$ \\

   	$\SDM_3=\frac{16+\cbox{yellow}{4\sqrt{2}+4\sqrt{10}}}{12}$ & 
   	$\SDM_3=\frac{16+\cbox{yellow}{3\sqrt{2}+5\sqrt{10}}}{12}$ \\

    \bottomrule
  \end{tabular}
  \caption{The Simplexwise Distance Distributions from Definition~\ref{dfn:SDD} for the 4-point clouds $T,K\subset\R^2$ in Fig.~\ref{fig:4-point_clouds}. 
The symbol $\times 2$ indicates a doubled $\RDD$.
The three bottom rows show coordinates of $\SDM(C;2,1)\in\R^3$ from Definition~\ref{dfn:SDM} for $h=2$, $l=1$ and both $C=T,K$.
Different elements are \hl{highlighted}.
}
\label{tab:SDD+TK}
\end{table}

Some of the $\binom{m}{h}$ $\RDD$s in $\SDD(C;h)$ can be identical as in Example~\ref{exa:SDD}.
If we collapse any $l>1$ identical $\RDD$s into a single $\RDD$ with the \emph{weight} $l/\binom{m}{h}$, $\SDD$ can be considered as a weighted probability distribution 
of $\RDD$s.
\smallskip

The $m-h$ permutable columns of the matrix $R(C;A)$ in $\RDD$ from Definition~\ref{dfn:RDD} can be interpreted as $m-h$ unlabeled points in $\R^h$.
Since any isometry is bijective, the simplest metric respecting bijections is the bottleneck distance, which is also called the Wasserstein distance $W_{\infty}$. 

\begin{dfn}[bottleneck distance $W_{\infty}$]
\label{dfn:bottleneck}
For any vector $v=(v_1,\dots,v_n)\in \R^n$, the \emph{Minkowski} norm is $||v||_{\infty}=\max\limits_{i=1,\dots,n}|v_i|$.
For any vectors or matrices $N,N'$ of the same size, the \emph{Minkowski} distance is $L_{\infty}(N,N')=\max\limits_{i,j}|N_{ij}-N'_{ij}|$.
For clouds $C,C'\subset\R^n$ of $m$ unlabeled points, 
the \emph{bottleneck distance} $W_{\infty}(C,C')=\inf\limits_{g:C\to C'} \sup\limits_{p\in C}||p-g(p)||_{\infty}$ is minimized over all bijections $g:C\to C'$.
\end{dfn}

\begin{lem}[the max metric $M_{\infty}$ on $\RDD$s]
\label{lem:RDD+metric}
For any $m$-point clouds and ordered $h$-point subsets $A\subset C$ and $A'\subset C'$, set $d(\xi)=\max\{L_{\infty}(\xi(D(A)),D(A')),W_{\infty}(\xi(R(C;A)),R(C';A'))\}$ for a permutation $\xi\in S_h$ on $h$ points. 
Then the max metric $M_{\infty}(\RDD(C;A),\RDD(C';A'))=\min\limits_{\xi\in S_h}d(\xi)$
satisfies all metric axioms on $\RDD$s from Definition~\ref{dfn:RDD} and can be computed in time $O(h!(h^2 +m^{1.5}\log^h m) )$. 
\end{lem}

We will use only $h=n$ for Euclidean space $\R^n$, so the factor $h!$ in Lemma~\ref{lem:RDD+metric} is practically small for $n=2,3$.
\smallskip

For $h=1$ and a 1-point subset $A\subset C$, the matrix $D(A)$ is empty, so $d(\xi)=W_{\infty}(\xi(R(C;A)),R(C';A'))$.
The metric $M_{\infty}$ on $\RDD$s will be used for intermediate costs to get metrics between two unordered collections of $\RDD$s by using standard Definitions~\ref{dfn:LAC} and~\ref{dfn:EMD} below.  

\begin{dfn}[Linear Assignment Cost LAC {\cite{fredman1987fibonacci}}]
\label{dfn:LAC}
For any $k\times k$ matrix of costs $c(i,j)\geq 0$, $i,j\in\{1,\dots,k\}$, the \emph{Linear Assignment Cost}   
$\LAC=\frac{1}{k}\min\limits_{g}\sum\limits_{i=1}^k c(i,g(i))$ is minimized for all bijections $g$ on the indices $1,\dots,k$.
\end{dfn}

The normalization factor $\frac{1}{k}$ in $\LAC$ makes this metric better comparable with $\EMD$ whose weights sum up to 1.
  
\begin{dfn}[Earth Mover's Distance on distributions]
\label{dfn:EMD}
Let $B=\{B_1,\dots,B_k\}$ be a finite unordered set of objects with weights $w(B_i)$, $i=1,\dots,k$.
Consider another set $D=\{D_1,\dots,D_l\}$ with weights $w(D_j)$, $j=1,\dots,l$.
Assume that a distance between $B_i,D_j$ is measured by a metric $d(B_i,D_j)$.
A \emph{flow} from $B$ to $D$ is a $k\times l$ matrix  whose entry $f_{ij}\in[0,1]$ represents a partial \emph{flow} from an object $B_i$ to $D_j$.
The \emph{Earth Mover's Distance} \cite{rubner2000earth} is the minimum of
$\EMD(B,D)=\sum\limits_{i=1}^{k} \sum\limits_{j=1}^{l} f_{ij} d(B_i,D_j)$ over $f_{ij}\in[0,1]$ subject to 
$\sum\limits_{j=1}^{l} f_{ij}\leq w(B_i)$ for $i=1,\dots,k$, 
$\sum\limits_{i=1}^{k} f_{ij}\leq w(D_j)$ for $j=1,\dots,l$, and
$\sum\limits_{i=1}^{k}\sum\limits_{j=1}^{l} f_{ij}=1$.
\end{dfn}

The first condition $\sum\limits_{j=1}^{l} f_{ij}\leq w(B_i)$ means that not more than the weight $w(B_i)$ of the object $B_i$ `flows' into all $D_j$ via the flows $f_{ij}$, $j=1,\dots,l$. 
The second condition $\sum\limits_{i=1}^{k} f_{ij}\leq w(D_j)$ means that all flows $f_{ij}$ from $B_i$ for $i=1,\dots,k$ `flow' to $D_j$ up to its weight $w(D_j)$.
The last condition
$\sum\limits_{i=1}^{k}\sum\limits_{j=1}^{l} f_{ij}=1$ forces all $B_i$ to collectively `flow' into all $D_j$.  
$\LAC$ \cite{fredman1987fibonacci} and $\EMD$ \cite{rubner2000earth} can be computed in a near cubic time in the sizes of given sets of objects. 
\smallskip

Theorems~\ref{thm:SDD}(c) and~\ref{thm:SCD} will extend $O(m^{1.5}\log^n m)$ algorithms for fixed clouds of $m$ unlabeled points in \cite[Theorem~6.5]{efrat2001geometry} to the harder case of isometry classes but keep the polynomial time in $m$ for a fixed dimension $n$.
All complexities are for a random-access machine (RAM) model.

\begin{thm}[invariance and continuity of $\SDD$s]
\label{thm:SDD}
\textbf{(a)}
For $h\geq 1$ and any cloud $C$ of $m$ unlabeled points in a metric space, $\SDD(C;h)$ is an isometry invariant, which can be computed in time $O(m^{h+1}/(h-1)!)$.
For any $l\geq 1$, the invariant $\SDM(C;h,l)\in\R^{m+\frac{h(h-3)}{2}}$ has the same time.
\smallskip

For any $m$-point clouds $C,C'$ in their own metric spaces and $h\geq 1$, let the Simplexwise Distance Distributions $\SDD(C;h)$ and $\SDD(C';h)$ consist of $k=\binom{m}{h}$ $\RDD$s with equal weights $\frac{1}{k}$ without collapsing identical $\RDD$s.
\smallskip

\noindent
\textbf{(b)}
Using the $k\times k$ matrix of costs computed by the metric $M_{\infty}$ between $\RDD$s from $\SDD(C;h)$ and $\SDD(C';h)$, 
the Linear Assignment Cost $\LAC$ from Definition~\ref{dfn:LAC} satisfies all metric axioms on $\SDD$s and can be computed in time $O(h!(h^2 +m^{1.5}\log^h m)k^2 + k^3\log k)$.
\smallskip

\noindent
\textbf{(c)}
Let $\SDD(C;h)$ and $\SDD(C';h)$ have a maximum size $l\leq k$ after collapsing identical $\RDD$s. Then $\EMD$ from Definition~\ref{dfn:EMD} satisfies all metric axioms  on $\SDD$s and is computed in time $O(h!(h^2 +m^{1.5}\log^h m) l^2 +l^3\log l)$.
\smallskip

\noindent
\textbf{(d)}
Let $C'$ be obtained from $C$ by perturbing each point within its $\ep$-neighborhood.
For any $h\geq 1$, $\SDD(C;h)$ changes by at most $2\ep$ in the $\LAC$ and $\EMD$ metrics.
The lower bound holds: $\EMD\big(\SDD(C;h),\SDD(C';h)\big)\geq|\SDM(C;h,1)-\SDM(C';h,1)|_\infty$.
\end{thm}

Theorem~\ref{thm:SDD}(d) substantially generalizes the fact that perturbing two points in their $\ep$-neighborhoods changes the Euclidean distance between these points by at most $2\ep$. 
\smallskip

We conjecture that $\SDD(C;h)$ is a complete isometry invariant of a cloud $C\subset\R^n$ for some $h\geq n-1$.
\cite[section~4]{kurlin2023simplexwise} shows that $\SDD(C;2)$ distinguished all infinitely many known pairs \cite[Fig.~S4]{pozdnyakov2020incompleteness} of non-isometric $m$-point clouds $C,C'\subset\R^3$ with identical $\PDD(C)=\SDD(C;1)$. 

\section{Simplexwise Centered Distribution (SCD)}
\label{sec:SCD}

While all constructions of section~\ref{sec:SDD} hold in any metric space, this section develops faster continuous metrics for complete isometry invariants of unlabeled clouds in $\R^n$.
\smallskip

The Euclidean structure of $\R^n$ allows us to translate the \emph{center of mass} $\dfrac{1}{m}\sum\limits_{p\in C} p$ of a given $m$-point cloud $C\subset\R^n$ to the origin $0\in\R^n$.
Then Problem~\ref{pro:isometry} reduces to only rotations around $0$ from the orthogonal group $\Or(\R^n)$.
\smallskip

Though the center of mass is  uniquely determined by any cloud $C\subset\R^n$ of unlabeled points, real applications may offer one or  several labeled points of $C$ that substantially speed up metrics on invariants.
For example, an atomic neighborhood in a solid material is a cloud $C\subset\R^3$ of atoms around a central atom, which may not be the center of mass of $C$, but is suitable for all methods below.
\smallskip

This section studies metrics on complete invariants of $C\subset\R^n$ up to rotations around the origin $0\in\R^n$, which may or may not belong to $C$ or be its center of mass.
\smallskip

For any subset $A=\{p_1,\dots,p_{n-1}\}\subset C$, the distance matrix $D(A\cup\{0\})$ from Definition~\ref{dfn:RDD} has size $(n-1)\times(n-1)$ and its last column can be chosen to include 
the distances from $n-1$ points of $A$ to the origin at $0\in\R^n$.
\smallskip

Any $n$ vectors $v_1,\dots,v_n\in\R^n$ can be written as columns in the $n\times n$ matrix whose determinant has a \emph{sign} $\pm 1$ or $0$ if $v_1,\dots,v_n$ are linearly dependent.
Any permutation $\xi\in S_n$ on indices $1,\dots,n$ is a composition of some $t$ transpositions $i\lra j$ and has $\sign(\xi)=(-1)^t$. 

\begin{dfn}[Simplexwise Centered Distribution $\SCD$]
\label{dfn:SCD}
Let $C\subset\R^n$ be any cloud of $m$ unlabeled points.
For any ordered subset $A$ of points $p_1,\dots,p_{n-1}\in C$, the matrix $R(C;A)$ from Definition~\ref{dfn:RDD} has a column of Euclidean distances $|q-p_1|,\dots,|q-p_{n-1}|$.
At the bottom of this column, add the distance $|q-0|$ to the origin and the sign of the determinant of the $n\times n$ matrix consisting of the vectors $q-p_1,\dots,q-p_{n-1},q$.  
The resulting $(n+1)\times(m-n+1)$-matrix with signs in the bottom $(n+1)$-st row is the \emph{oriented relative distance} matrix $M(C;A\cup\{0\})$.
\smallskip

Any permutation $\xi\in S_{n-1}$ of $n-1$ points of $A$ acts on $D(A)$, permutes the first $n-1$ rows of $M(C;A\cup\{0\})$ and multiplies every sign in the $(n+1)$-st row by $\sign(\xi)$.
\smallskip

The \emph{Oriented Centered Distribution} $\OCD(C;A)$ is the equivalence class of pairs $[D(A\cup\{0\}),M(C;A\cup\{0\})]$ considered up to permutations $\xi\in S_{n-1}$ of points of $A$.
\smallskip

The \emph{Simplexwise Centered Distribution} $\SCD(C)$ is the unordered set of the distributions $\OCD(C;A)$ for all $\binom{m}{n-1}$ unordered $(n-1)$-point subsets $A\subset C$.
The mirror image $\mSCD(C)$ is obtained from $\SCD(C)$ by reversing signs.
\end{dfn}

Definition~\ref{dfn:SCD} needs no permutations for any $C\subset\R^2$ as $n-1=1$.
Columns of $M(C;A\cup\{0\})$ can be lexicographically ordered without affecting the metric in Lemma~\ref{lem:OCD+metric}.
Some of the $\binom{m}{n-1}$ $\OCD$s in $\SCD(C)$ can be identical as in Example~\ref{exa:SCD}(b).
If we collapse any $l>1$ identical $\OCD$s into a single $\OCD$ with the \emph{weight} $l/\binom{m}{h}$, $\SCD$ can be considered as a weighted probability distribution 
of $\OCD$s.

\begin{exa}[$\SCD$ for clouds in Fig.~\ref{fig:triangular_clouds}]
\label{exa:SCD}
\textbf{(a)}
Let $R\subset\R^2$ consist of the vertices $p_1=(0,0)$, $p_2=(4,0)$, $p_3=(0,3)$ of the right-angled triangle in Fig.~\ref{fig:triangular_clouds}~(middle).
Though $p_1=(0,0)$ is included in $R$ and is not its center of mass,
$\SCD(R)$ still makes sense.
In
$\OCD(R;p_1)=[0,\left( \begin{array}{cc} 
4 & 3 \\
4 & 3 \\
0 & 0
\end{array}\right) ]$,
the matrix $D(\{p_1,0\})$ is $|p_1-0|=0$,
the top row has $|p_2-p_1|=4$, $|p_3-p_1|=3$.
In $\OCD(R;p_2)=[4,\left( \begin{array}{cc} 
4 & 5 \\
0 & 3 \\
0 & -
\end{array}\right) ]$,
the first row has $|p_1-p_2|=4$, $|p_3-p_2|=5$,
the second row has $|p_1-0|=0$, $|p_3-0|=3$,
$\det\mat{-4}{0}{3}{3}<0$.
In $\OCD(R;p_3)=[3,\left( \begin{array}{cc} 
3 & 5 \\
0 & 4 \\
0 & +
\end{array}\right) ]$,
the first row has $|p_1-p_3|=3$, $|p_2-p_3|=5$,
the second row has $|p_1-0|=0$, $|p_2-0|=4$,
$\det\mat{4}{4}{-3}{0}>0$.
So $\SCD(R)$ consists of the three Oriented Centered Distributions $\OCD$s above.
\smallskip

If we reflect $R$ with respect to the $x$-axis, the new cloud $\bar R$ of the points $p_1,p_2,\bar p_3=(0,-3)$ has $\SCD(\bar R)=\mSCD(R)$ with
$\OCD(\bar R;p_1)=\OCD(R)$,
$\OCD(\bar R;p_2)=[4,\left( \begin{array}{cc} 
4 & 5 \\
0 & 3 \\
0 & +
\end{array}\right) ]$,
$\OCD(R;\bar p_3)=[3,\left( \begin{array}{cc} 
3 & 5 \\
0 & 4 \\
0 & -
\end{array}\right) ]$ whose signs changed under reflection, so $\SCD(R)\neq\SCD(\bar R)$.
\medskip

\noindent
\textbf{(b)}
Let $S\subset\R^2$ consist of $m=4$ points $(\pm 1,0),(0,\pm 1)$ that are vertices of the square in Fig.~\ref{fig:triangular_clouds}~(right).
The center of mass is $0\in\R^2$ and has a distance $1$ to each point of $S$.
\smallskip

For each 1-point subset $A=\{p\}\subset S$, the distance matrix $D(A\cup\{0\})$ on two points is the single number $1$.
The matrix $M(S;A\cup\{0\})$ has $m-n+1=3$ columns.
For $p_1=(1,0)$, we have   
$M(S;\vect{p_1}{0})=\left(\begin{array}{ccc} 
\sqrt{2} & \sqrt{2} & 2 \\
1 & 1 & 1 \\
- & + & 0
\end{array}\right)$, where the columns are ordered according to
$p_2=(0,-1)$, $p_3=(0,1)$, $p_4=(-1,0)$ in Fig.~\ref{fig:triangular_clouds}~(right).
The sign in the bottom right corner is 0 because the points $p_1,0,p_4$ are in a straight line. 
Due to the rotational symmetry, $M(S;\{p_i,0\})$ is independent of $i=1,2,3,4$.
So $\SCD(S)$ can be considered as one $\OCD=[1,M(S;\vect{p_1}{0})]$ of weight 1. 
\end{exa}

Example~\ref{exa:SCD}(b) illustrates the key discontinuity challenge:
if $p_4=(-1,0)$ is perturbed, the corresponding sign can discontinuously change to $+1$ or $-1$.
To get a continuous metric on $\OCD$s, we will multiply each sign by 
a continuous \emph{strength} function that vanishes for any zero sign.

\begin{dfn}[\emph{strength} $\si(A)$ of a simplex]
\label{dfn:strength}
For a set $A$ of $n+1$ points $q=p_0,p_1,\dots,p_n$ in $\R^n$,
let $p(A)=\frac{1}{2}\sum\limits_{i\neq j}^{n+1}|p_i-p_j|$ be half of the sum of all pairwise distances.
Let $V(A)$ denote the volume the $n$-dimensional simplex on the set $A$.
Define the \emph{strength} $\si(A)=V^2(A)/p^{2n-1}(A)$.
\smallskip

For $n=2$ and a triangle $A$ with sides $a,b,c$ in $\R^2$, Heron's formula 
gives $\si(A)=\dfrac{(p-a)(p-b)(p-c)}{p^2}$, $p=\dfrac{a+b+c}{2}=p(A)$ is the half-perimeter of $A$.
\end{dfn}

For $n=1$ and a set $A={p_0,p_1}\subset\R$, the volume is $V(A)=|p_0-p_1|=2p(A)$, so $\si(A)=
2|p_0-p_1|$.
\smallskip
 
The strength $\si(A)$ depends only on the distance matrix $D(A)$ from Definition~\ref{dfn:RDD}, so the notation $\si(A)$ is used only for brevity.
In any $\R^n$, the squared volume $V^2(A)$ is expressed by the Cayley-Menger determinant \cite{sippl1986cayley} in pairwise distances between points of $A$.
Importantly, the strength $\si(A)$ vanishes when the simplex on a set $A$ degenerates. 
\smallskip

Theorem~\ref{thm:SCD} will need the continuity of $s\si(A)$, when a sign $s\in\{\pm 1\}$ from a bottom row of $\ORD$ discontinuously changes while passing through a degenerate set $A$.
The proof of the continuity of $\si(A)$ in Theorem~\ref{thm:strength} gives an explicit upper bound for a Lipschitz constant $c_n$ below. 

\begin{thm}[Lipschitz continuity of the strength $\si$]
\label{thm:strength}
Let a cloud $A'$ be obtained from another $(n+1)$-point cloud $A\subset\R^n$ by perturbing every point within its $\ep$-neighborhood.
The strength $\si(A)$ from Definition~\ref{dfn:strength} is Lipschitz continuous so that $|\si(A')-\si(A)|\leq 2\ep c_n$ for a constant $c_n$.
\end{thm}

\begin{figure}[h!]
\centering
\includegraphics[width=\linewidth]{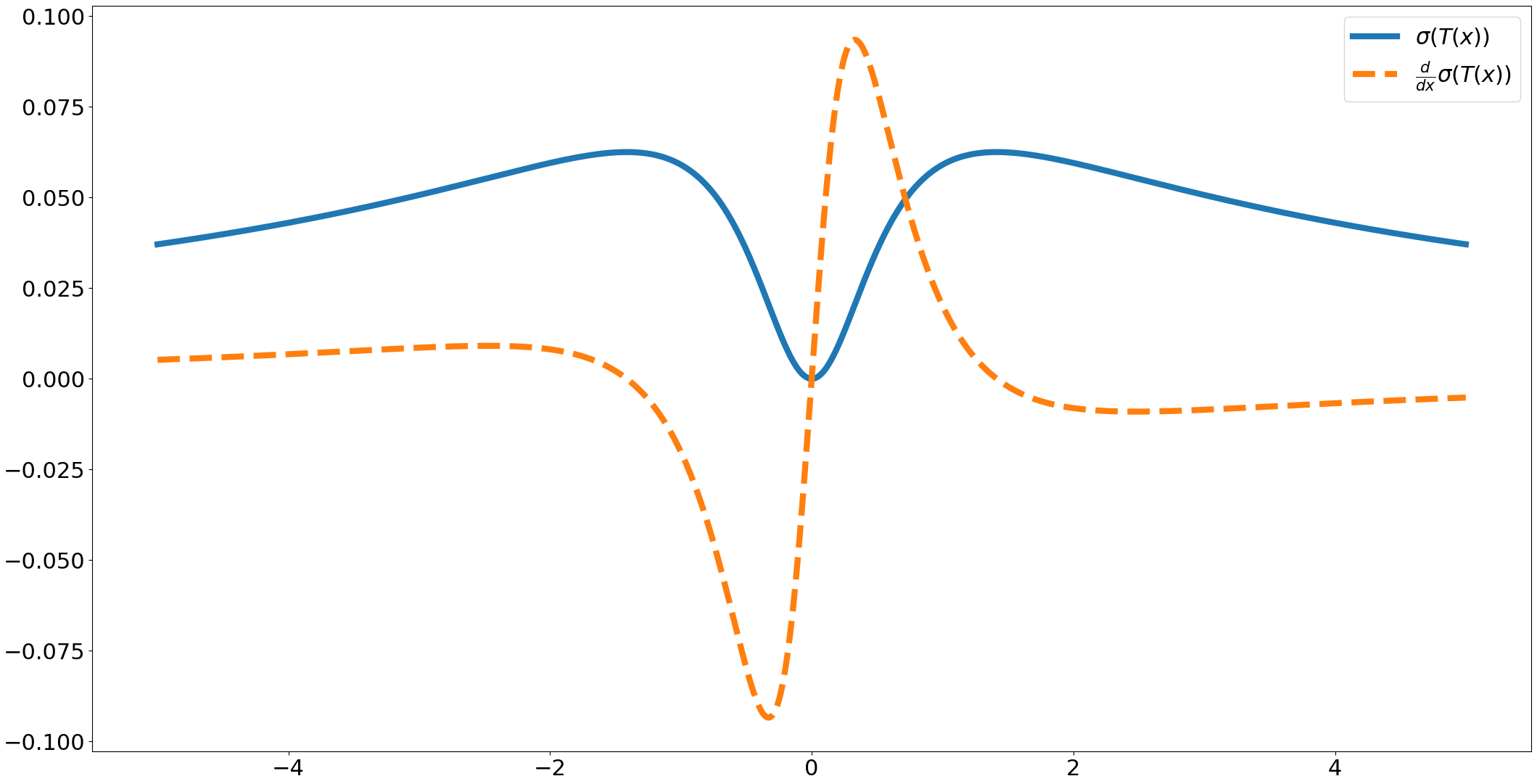}

\includegraphics[width=\linewidth]{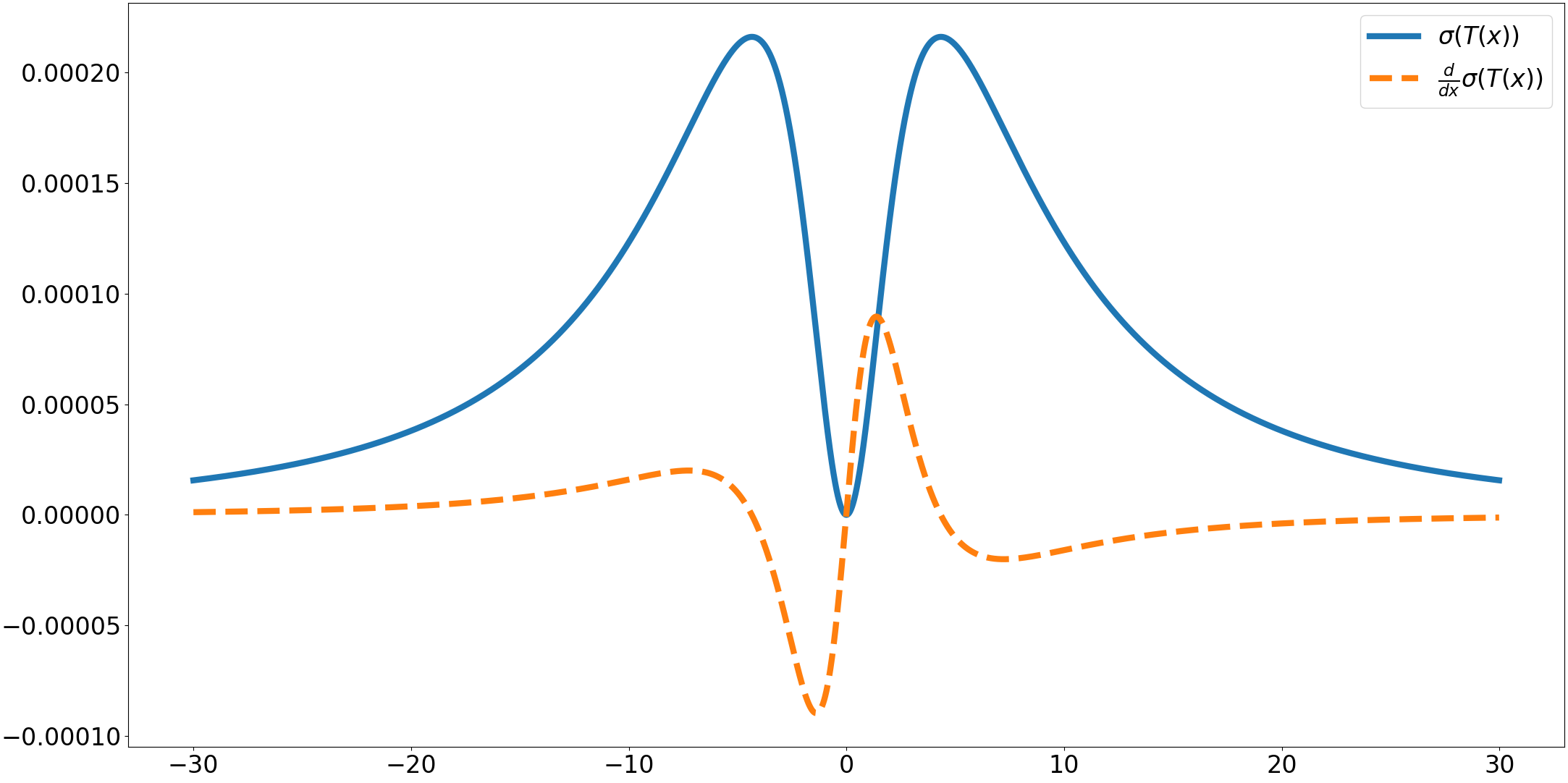}
\caption{
The strength $\si$ (solid curve) and its derivative $\pd{\si}{x}$ (dashed curve) in the $x$-coordinate of a point from $A$ were averaged over 3000 random triangles (top) and tetrahedra (bottom).}
\label{fig:strength}
\end{figure}

\begin{exa}[strength $\si(A)$ and its upper bounds]
\label{exa:strength}
\cite[Theorem~4.2]{kurlin2023strength} proves upper bounds for the Lipschitz constant of the strength: $c_2=2\sqrt{3}$, $c_3\approx 0.43$, $c_4\approx 0.01$, which quickly tend to 0 due to the `curse of dimensionality'. 
The plots in Fig.~\ref{fig:strength} illustrate that the strength $\si()$ behaves smoothly in the $x$-coordinate of a vertex and its derivative $|\pd{\si}{x}|$ is much smaller than the proved bounds $c_n$ above. 
\end{exa}

The strength $\si(A)$ from Definition~\ref{dfn:strength} will take care of extra signs in ORDs and allows us to prove the analogue of Lemma~\ref{lem:RDD+metric} for a similar time complexity with $h=n$.  

\begin{lem}[metric on $\OCD$s]
\label{lem:OCD+metric}
Using the strength $\si$ from Definition~\ref{dfn:strength}, we 
consider the bottleneck distance $W_{\infty}$ on the set of permutable $m-n+1$ columns of $M(C;A\cup\{0\})$ as on the set of $m-n+1$ unlabeled points $\left(v,\dfrac{s}{c_{n}}\si(A\cup\{0,q\})\right)\in\R^{n+1}$.
For another $\OCD'=[D(A'\cup\{0\});M(C';A'\cup\{0\})]$ and any permutation $\xi\in S_{n-1}$ of indices $1,\dots,n-1$ acting on $D(A\cup\{0\})$ and the first $n-1$ rows of $M(C;A\cup\{0\})$, set $d_o(\xi)=\max\{L,W\}$, 
$$\text{ where }L=L_{\infty}\Big(\xi(D(A\cup\{0\})),D(A'\cup\{0\})\Big),$$
$$W=W_{\infty}\Big(\xi(M(C;A\cup\{0\})),M(C';A'\cup\{0\})\Big).$$
Then $M_\infty(\OCD,\OCD')=\min\limits_{\xi\in S_{n-1}} d_o(\xi)$
satisfies all metric axioms on Oriented Centered Distributions ($\OCD$s) and is computed in time $O((n-1)!(n^2+m^{1.5}\log^{n} m))$.
\end{lem}

The coefficient $\frac{1}{c_{n}}$ 
normalizes the Lipschitz constant $c_{n}$ of $\si$ to $1$ in line with changes of distances by at most $2\ep$ when points are perturbed within their $\ep$-neighborhoods.
An equality $\SCD(C)=\SCD(C')$ is interpreted as a bijection between unordered sets $\SCD(C)\to\SCD(C')$ matching all $\OCD$s, which is best detected by checking if metrics in Theorem~\ref{thm:SCD} between these $\SCD$s is 0.

\begin{thm}[completeness and continuity of $\SCD$]
\label{thm:SCD}
\textbf{(a)}
The Simplexwise Centered Distribution $\SCD(C)$ in Definition~\ref{dfn:SCD} is a complete isometry invariant of clouds $C\subset\R^n$ of $m$ unlabeled points with a center of mass at the origin $0\in\R^n$, and can be computed in time $O(m^n/(n-4)!)$.
\smallskip

So any clouds $C,C'\subset\R^n$ are related by rigid motion (isometry, respectively) \emph{if and only if} 
$\SCD(C)=\SCD(C')$ ($\SCD(C)$ equals $\SCD(C')$ or its mirror image $\mSCD(C')$, respectively).
For any $m$-point clouds $C,C'\subset\R^n$, let 
$\SCD(C)$ and $\SCD(C')$ consist of $k=\binom{m}{n-1}$ $\OCD$s.
\medskip

\noindent
\textbf{(b)}
For the $k\times k$ matrix of costs computed by the metric $M_{\infty}$ between $\OCD$s in $\SCD(C)$ and $\SCD(C')$, 
$\LAC$ from Definition~\ref{dfn:LAC} satisfies all metric axioms on $\SCD$s and needs time $O((n-1)!(n^2+m^{1.5}\log^{n} m)k^2+ k^3\log k)$.
\medskip

\noindent
\textbf{(c)}
Let $\SCD$s have a maximum size $l\leq k$ after collapsing identical $\OCD$s. Then $\EMD$ from Definition~\ref{dfn:EMD} satisfies all metric axioms  on $\SCD$s and can be computed in time $O((n-1)!(n^2 +m^{1.5}\log^{n} m) l^2 +l^3\log l)$.
\medskip

\noindent
\textbf{(d)}
Let $C'$ be obtained from a cloud $C\subset\R^n$ by perturbing each point within its $\ep$-neighborhood.
Then $\SCD(C)$ changes by at most $2\ep$ in the $\LAC$ and $\EMD$ metrics.
\end{thm}

If we estimate $l\leq k=\binom{m}{n-1}=m(m-1)\dots(m-n+2)/n!$ as $O(m^{n-1}/n!)$, Theorem~\ref{thm:SCD}(b,c) gives time 
$O(n(m^{n-1}/n!)^3\log m)$ for metrics on $\SCD$s, which is $O(m^3\log m)$ for $n=2$, and $O(m^6\log m)$ for $n=3$. 
\smallskip

Though the above time estimates are very rough upper bounds, 
the time $O(m^3\log m)$ in $\R^2$ is faster than the only past time $O(m^5\log m)$ for comparing $m$-point clouds by the Hausdorff distance minimized over isometries \cite{chew1997geometric}.

\begin{dfn}[Centered Distance Moments $\CDM$]
\label{dfn:CDM}
For any $m$-point cloud $C\subset\R^n$,
 let $A\subset C$ be a subset of $n-1$ unordered points.
The \emph{Centered Interpoint Distance} list $\CID(A)$ is the increasing list of all $\frac{(n-1)(n-2)}{2}$ pairwise distances between points of $A$, followed by $n-1$ increasing distances from $A$ to the origin $0$. 
For each column of the $(n+1)\times(m-n+1)$ matrix $M(C;A\cup\{0\})$ in Definition~\ref{dfn:SCD}, compute the average of the first $n-1$ distances.
Write these averages in increasing order, append the list of increasing distances $|q-0|$ from the $n$-th row of $M(C;A\cup\{0\})$, and also append the vector of increasing values of $\dfrac{s}{c_n}\si(A\cup\{0\})$ taking signs $s$ from the $(n+1)$-st row of $M(C;A\cup\{0\})$.
Let $\vec M(C;A)\in\R^{3(m-n+1)}$ be the final vector.
\smallskip

The pair $[\CID(A);\vec M(C;A)]$ is the \emph{Average Centered Vector} $\ACV(C;A)$ considered as a vector of length $\frac{n(n-1)}{2}+3(m-n+1)$.
The unordered set of $\ACV(C;A)$ for all $\binom{m}{n-1}$ unordered subsets $A\subset C$ is the Average Centered Distribution $\ACD(C)$.
The \emph{Centered Distance Moment} $\CDM(C;l)$ is the $l$-th (standardized for $l\geq 3$) moment of $\ACD(C)$ considered as a probability distribution of $\binom{m}{n-1}$ vectors, separately for each coordinate.
\end{dfn}

\begin{exa}[$\CDM$ for clouds in Fig.~\ref{fig:triangular_clouds}]
\textbf{(a)}
For $n=2$ and the cloud $R\subset\R^2$ of $m=3$ vertices $p_1=(0,0)$, $p_2=(4,0)$, $p_3=(0,3)$ of the right-angled triangle in Fig.~\ref{fig:triangular_clouds}~(middle), we continue Example~\ref{exa:SCD}(a) and flatten $\OCD(R;p_1)=[0,\left( \begin{array}{cc} 
4 & 3 \\
4 & 3 \\
0 & 0
\end{array}\right) ]$ into the vector $\ACV(R;p_1)=[0;3,4;3,4;0,0]$ of length $\frac{n(n-1)}{2}+3(m-n+1)=7$, whose four parts ($1+2+2+2=7$) are in increasing order, similarly for $p_2,p_3$.
The Average Centered Distribution can be written as a $3\times 7$ matrix with unordered rows: $\ACD(R)=\left( \begin{array}{c|cc|cc|cc} 
0 & 3 & 4 & 3 & 4 & 0 & 0 \\
4 & 4 & 5 & 0 & 3 & 0 & -6/c_2 \\
3 & 3 & 5 & 0 & 4 & 0 & 6/c_2
\end{array}\right)$.
The area of the triangle on $R$ equals $6$ and can be normalized by $c_2=2\sqrt{3}$ to get $6/c_2=\sqrt{3}$, see \cite[section~4]{kurlin2023strength}.
The 1st moment is $\CDM(R;1)=\frac{1}{3}(7;10,14;3,11;0)$.
\medskip

\noindent
\textbf{(b)}
For $n=2$ and the cloud $S\subset\R^2$ of $m=4$ vertices 
of the square in Fig.~\ref{fig:triangular_clouds}~(right), 
Example~\ref{exa:SCD}(a) computed $\SCD(R)$ as one
$\OCD=[1,\left(\begin{array}{ccc} 
\sqrt{2} & \sqrt{2} & 2 \\
1 & 1 & 1 \\
- & + & 0
\end{array}\right)]$, which flattens to 
$\ACV=(1; \sqrt{2}, \sqrt{2}, 2; 1,1,1; -\frac{1}{2}, \frac{1}{2},0)=\ACD(S)=\CDM(S;1)\in\R^{10}$, where $\frac{1}{2}$ is the area of the triangle on the vertices $(0,0),(1,0),(0,1)$.
\end{exa}

\begin{cor}[time for continuous metrics on $\CDM$s]
\label{cor:CDM}
For any cloud $C\subset\R^n$ of $m$ unlabeled points,
the Centered Distance Moment $\CDM(C;l)$ in Definition~\ref{dfn:CDM} is computed in time $O(m^n/(n-4)!)$.
The metric $L_\infty$ on $\CDM$s needs $O(n^2+m)$ time and 
$\EMD\big(\SCD(C),\SCD(C')\big)\geq|\CDM(C;1)-\CDM(C';1)|_\infty$ holds.
\end{cor}

\section{Experiments and discussion of future work}
\label{sec:experiments}


This paper advocates a scientific approach to any data exemplified by Problem~\ref{pro:isometry}, where rigid motion on clouds can be replaced by another equivalence of other data.
The scientific principles such as axioms should be always respected.
Only the first coincidence axiom in (\ref{pro:isometry}b) guarantees no duplicate data.
If the triangle inequality fails with any additive error, results of clustering can be pre-determined \cite{rass2022metricizing}.  
\smallskip

The notorious $m!$ challenge of $m$ \emph{unlabeled} points in  Problem~\ref{pro:isometry} was solved in $\R^n$ by Theorem~\ref{thm:SCD}, also up to rigid motion by using the novel \emph{strength} of a simplex to smooth signs of determinants due to hard Theorem~\ref{thm:strength}. 
\smallskip

The results above sufficiently justify re-focusing future efforts from experimental attempts at Problem~\ref{pro:isometry} to higher level tasks such as predicting properties of rigid objects, e.g. crystalline materials, using the complete invariants with \emph{no false negatives} and \emph{no false positives} for all possible data since \emph{no experiments} can beat the proved 100\% guarantee.
\smallskip

To tackle the limitation of comparing only clouds having a fixed number $m$ of points, the Earth Mover's Distance ($\EMD$) continuously can compare any distributions ($\SDD$ or $\SCD$) of different sizes.
Using $\EMD$ instead of the bottleneck distance $W_{\infty}$ on $m-h$ (or $m-n+1$) columns of matrices in Definitions~\ref{dfn:SDD} and~\ref{dfn:SCD} increases a time from $O(m^{1.5}\log m)$ to $O(m^3\log m)$ but the total time remains the same due to a near cubic time 
in the last step.
\smallskip

The running time in real applications is smaller for several reasons.
First, the shape (isometry class) of any rigid body in $\R^3$ is determined by only $m=4$ labeled points in general position.
Even when points are unlabeled, dozens of corners or feature points suffice to represent a rigid shape well enough. 
Second, the key size $l$ (number of distinct Oriented Centered Distributions) in Theorem~\ref{thm:SCD} is often smaller than $m$, especially for symmetric objects, see $l=1<m=4$ in Example~\ref{exa:SCD}.
The $\SCD$ invariants are on top of others due to their completeness and continuity.
\smallskip

The past work \cite{widdowson2022average, widdowson2022resolving} used the simpler Pointwise Distance Distribution (PDD) to complete 200B+ pairwise comparisons of all 660K+ periodic crystals in the world's largest database of real materials.
This experiment took only a couple of days on a modest desktop and established the \emph{Crystal Isometry Principle} saying that any real periodic crystal is uniquely determined by the geometry of its atomic centers without chemical elements.
So the type of any atom is provably reconstructable from distances to atomic neighbors.
\smallskip

\begin{figure}[h!]
\centering
\includegraphics[width=\columnwidth]{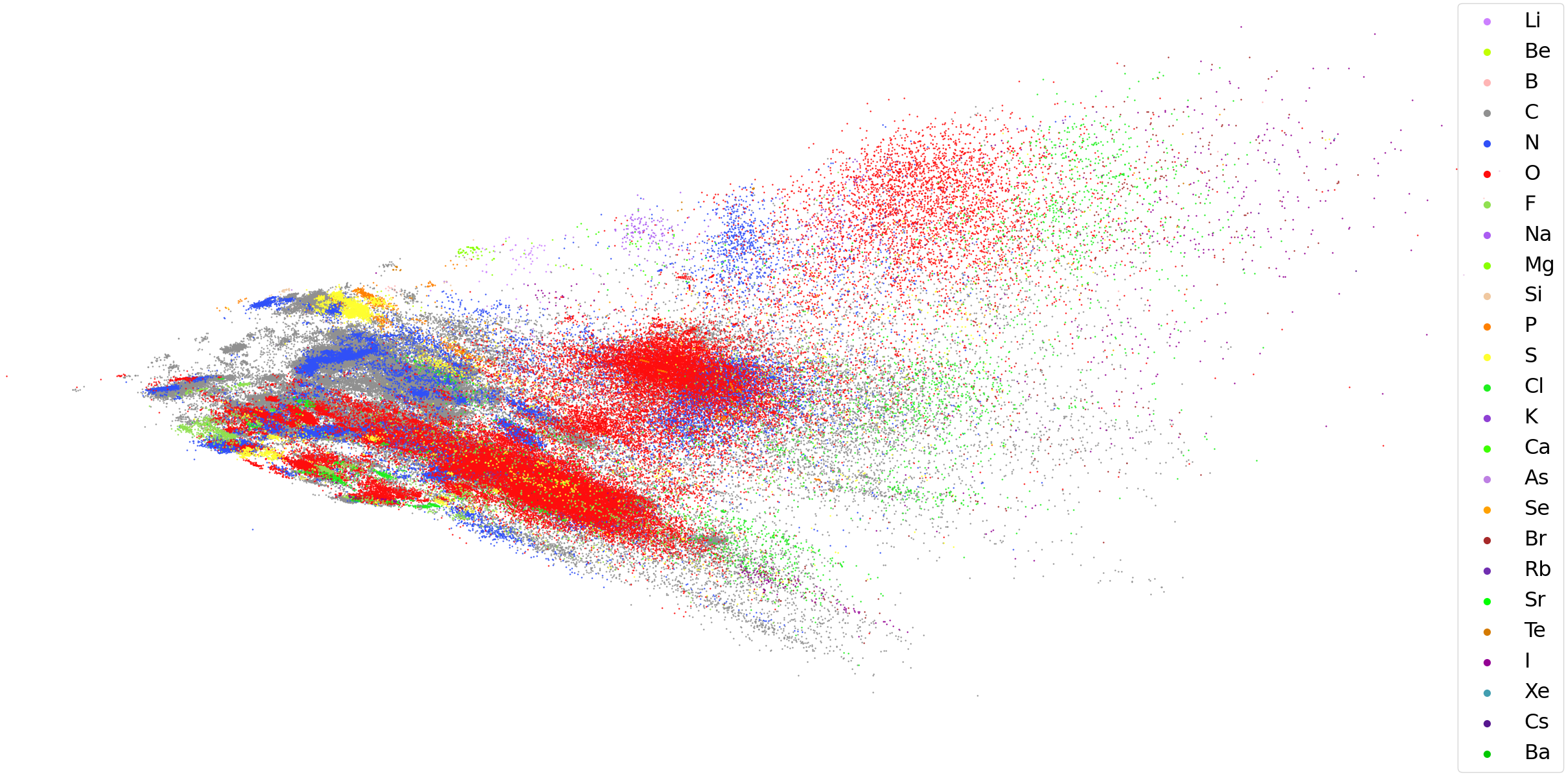}
\caption{Two principal directions of $\SDV$s for all 300K+ atomic clouds from all 10K+ drugs in the CSD, colored by 25 elements.}
\label{fig:CSD_drugs_25atom_clouds_OPD5_PCA2}
\end{figure}

The new invariants allow us to go deeper and compare atomic clouds from higher level periodic crystals.
Fig.~\ref{fig:CSD_drugs_25atom_clouds_OPD5_PCA2} visualizes all 300K+ atomic clouds extracted from all 10K+ crystalline drugs in the Cambridge Structural Database (CSD) by using $\SDV$ invariants for $5+1$ atoms including the central one.
Future maps will use stronger invariants. 

\smallskip

This research was supported by the Royal Academy Engineering fellowship IF2122/186, and EPSRC grants EP/R018472/1, EP/X018474/1.
We thank all members of the Data Science Theory and Applications group at Liverpool (UK) and all reviewers for their helpful suggestions.

{\small
\bibliographystyle{ieee_fullname}
\bibliography{CVPR2023}

\begin{thebibliography}{10}\itemsep=-1pt

\bibitem{akhtar2018threat}
Naveed Akhtar and Ajmal Mian.
\newblock Threat of adversarial attacks on deep learning in computer vision: A
  survey.
\newblock {\em IEEE Access}, 6:14410--14430, 2018.

\bibitem{anosova2021introduction}
Olga Anosova and Vitaliy Kurlin.
\newblock Introduction to periodic geometry and topology.
\newblock {\em arXiv:2103.02749}.

\bibitem{anosova2021isometry}
Olga Anosova and Vitaliy Kurlin.
\newblock An isometry classification of periodic point sets.
\newblock In {\em Proceedings of Discrete Geometry and Mathematical
  Morphology}, 2021.

\bibitem{anosova2022algorithms}
Olga Anosova and Vitaliy Kurlin.
\newblock Algorithms for continuous metrics on periodic crystals.
\newblock {\em arxiv:2205.15298}, 2022.

\bibitem{anosova2022density}
Olga Anosova and Vitaliy Kurlin.
\newblock Density functions of periodic sequences.
\newblock {\em Lecture Notes in Computer Science (Proceedings of DGMM)}, 2022.

\bibitem{anosova2023density}
O Anosova and V Kurlin.
\newblock Density functions of periodic sequences of continuous events.
\newblock {\em arXiv:2301.05137}, 2023.

\bibitem{balasingham2022compact}
Jonathan Balasingham, Viktor Zamaraev, and Vitaliy Kurlin.
\newblock Compact graph representation of crystals using {P}ointwise {D}istance
  {D}istributions.
\newblock {\em arXiv:2212.11246}, 2022.

\bibitem{belongie2002shape}
Serge Belongie, Jitendra Malik, and Jan Puzicha.
\newblock Shape matching and object recognition using shape contexts.
\newblock {\em Transactions PAMI}, 24(4):509--522, 2002.

\bibitem{boutin2004reconstructing}
Mireille Boutin and Gregor Kemper.
\newblock On reconstructing n-point configurations from the distribution of
  distances or areas.
\newblock {\em Adv. Appl. Math.}, 32(4):709--735, 2004.

\bibitem{brass2000testing}
Peter Brass and Christian Knauer.
\newblock Testing the congruence of d-dimensional point sets.
\newblock In {\em SoCG}, pages 310--314, 2000.

\bibitem{brass2004testing}
Peter Brass and Christian Knauer.
\newblock Testing congruence and symmetry for general 3-dimensional objects.
\newblock {\em Computational Geometry}, 27(1):3--11, 2004.

\bibitem{bright2021welcome}
Matthew~J Bright, Andrew~I Cooper, and Vitaliy~A Kurlin.
\newblock Welcome to a continuous world of 3-dimensional lattices.
\newblock {\em arxiv:2109.11538}, 2021.

\bibitem{bright2023geographic}
Matthew~J Bright, Andrew~I Cooper, and Vitaliy~A Kurlin.
\newblock Geographic-style maps for 2-dimensional lattices.
\newblock {\em Acta Crystallographica Section A}, 79(1):1--13, 2023.

\bibitem{bronstein2021geometric}
Michael~M Bronstein, Joan Bruna, Taco Cohen, and Petar Veli{\v{c}}kovi{\'c}.
\newblock Geometric deep learning: grids, groups, graphs, geodesics, and
  gauges.
\newblock {\em arXiv:2104.13478}, 2021.

\bibitem{bronstein2017geometric}
Michael~M Bronstein, Joan Bruna, Yann LeCun, Arthur Szlam, and Pierre
  Vandergheynst.
\newblock Geometric deep learning: going beyond {E}uclidean data.
\newblock {\em IEEE Signal Processing Magazine}, 34(4):18--42, 2017.

\bibitem{chew1999geometric}
Paul Chew, Dorit Dor, Alon Efrat, and Klara Kedem.
\newblock Geometric pattern matching in d-dimensional space.
\newblock {\em Discrete \& Computational Geometry}, 21(2):257--274, 1999.

\bibitem{chew1997geometric}
Paul Chew, Michael Goodrich, Daniel Huttenlocher, Klara Kedem, Jon Kleinberg,
  and Dina Kravets.
\newblock Geometric pattern matching under {E}uclidean motion.
\newblock {\em Computational Geometry}, 7(1-2):113--124, 1997.

\bibitem{chew1992improvements}
Paul Chew and Klara Kedem.
\newblock Improvements on geometric pattern matching problems.
\newblock In {\em Scandinavian Workshop on Algorithm Theory}, pages 318--325,
  1992.

\bibitem{colbrook2022difficulty}
Matthew~J Colbrook, Vegard Antun, and Anders~C Hansen.
\newblock The difficulty of computing stable and accurate neural networks: On
  the barriers of deep learning and {S}male’s 18th problem.
\newblock {\em PNAS}, 119(12):e2107151119, 2022.

\bibitem{dong2018boosting}
Yinpeng Dong, Fangzhou Liao, Tianyu Pang, Hang Su, Jun Zhu, Xiaolin Hu, and
  Jianguo Li.
\newblock Boosting adversarial attacks with momentum.
\newblock In {\em Computer vision and pattern recognition}, pages 9185--9193,
  2018.

\bibitem{edelsbrunner2021density}
H Edelsbrunner, T Heiss, V Kurlin, P Smith, and M Wintraecken.
\newblock The density fingerprint of a periodic point set.
\newblock In {\em Proceedings of SoCG}, pages 32:1--32:16, 2021.

\bibitem{efrat2001geometry}
Alon Efrat, Alon Itai, and Matthew~J Katz.
\newblock Geometry helps in bottleneck matching and related problems.
\newblock {\em Algorithmica}, 31(1):1--28, 2001.

\bibitem{elkin2022new}
Yury Elkin.
\newblock New compressed cover tree for k-nearest neighbor search (phd thesis).
\newblock {\em arxiv:2205.10194}, 2022.

\bibitem{elkin2020mergegram}
Y. Elkin and V. Kurlin.
\newblock The mergegram of a dendrogram and its stability.
\newblock In {\em Proceedings of MFCS}, 2020.

\bibitem{elkin2021isometry}
Y. Elkin and V. Kurlin.
\newblock Isometry invariant shape recognition of projectively perturbed point
  clouds by the mergegram extending 0d persistence.
\newblock {\em Mathematics}, 9(17), 2021.

\bibitem{fredman1987fibonacci}
Michael~L Fredman and Robert~Endre Tarjan.
\newblock Fibonacci heaps and their uses in improved network optimization
  algorithms.
\newblock {\em Journal ACM}, 34:596--615, 1987.

\bibitem{goodrich1999approximate}
Michael~T Goodrich, Joseph~SB Mitchell, and Mark~W Orletsky.
\newblock Approximate geometric pattern matching under rigid motions.
\newblock {\em Transactions PAMI}, 21:371--379, 1999.

\bibitem{grigorescu2003distance}
Cosmin Grigorescu and Nicolai Petkov.
\newblock Distance sets for shape filters and shape recognition.
\newblock {\em IEEE transactions on image processing}, 12(10):1274--1286, 2003.

\bibitem{grinberg2019n}
Darij Grinberg and Peter~J Olver.
\newblock The n body matrix and its determinant.
\newblock {\em SIAM Journal on Applied Algebra and Geometry}, 3(1):67--86,
  2019.

\bibitem{guo2019simple}
Chuan Guo, Jacob Gardner, Yurong You, Andrew~Gordon Wilson, and Kilian
  Weinberger.
\newblock Simple black-box adversarial attacks.
\newblock In {\em ICML}, pages 2484--2493, 2019.

\bibitem{hausdorff1919dimension}
Felix Hausdorff.
\newblock Dimension und {\"a}u$\beta$eres ma$\beta$.
\newblock {\em Mathematische Annalen}, 79(2):157--179, 1919.

\bibitem{hordan2023complete}
Snir Hordan, Tal Amir, Steven~J Gortler, and Nadav Dym.
\newblock Complete neural networks for {E}uclidean graphs.
\newblock {\em arXiv:2301.13821}, 2023.

\bibitem{huttenlocher1993comparing}
Daniel Huttenlocher, Gregory Klanderman, and William Rucklidge.
\newblock Comparing images using the {H}ausdorff distance.
\newblock {\em Transactions PAMI}, 15:850--863, 1993.

\bibitem{keeping1995introduction}
Ernest~Sydney Keeping.
\newblock {\em Introduction to statistical inference}.
\newblock Courier Corporation, 1995.

\bibitem{kurlin2022complete}
Vitaliy Kurlin.
\newblock A complete isometry classification of 3-dimensional lattices.
\newblock {\em arxiv:2201.10543}, 2022.

\bibitem{kurlin2022computable}
Vitaliy Kurlin.
\newblock Computable complete invariants for finite clouds of unlabeled points
  under {E}uclidean isometry.
\newblock {\em arXiv:2207.08502}, 2022.

\bibitem{kurlin2022exactly}
Vitaliy Kurlin.
\newblock Exactly computable and continuous metrics on isometry classes of
  finite and 1-periodic sequences.
\newblock {\em arxiv:2205.04388}, 2022.

\bibitem{kurlin2022mathematics}
Vitaliy Kurlin.
\newblock Mathematics of 2-dimensional lattices.
\newblock {\em Foundations of Computational Mathematics}, pages 1--59, 2022.

\bibitem{kurlin2023simplexwise}
Vitaliy Kurlin.
\newblock Simplexwise distance distributions for finite spaces with metrics and
  measures.
\newblock {\em arXiv:2303.14161}, 2023.

\bibitem{kurlin2023strength}
Vitaliy Kurlin.
\newblock The strength of a simplex is the key to a continuous isometry
  classification of {E}uclidean clouds of unlabelled points.
\newblock {\em arXiv:2303.13486}, 2023.

\bibitem{laidlaw2019functional}
Cassidy Laidlaw and Soheil Feizi.
\newblock Functional adversarial attacks.
\newblock {\em Adv. Neural Inform. Proc. Systems}, 32, 2019.

\bibitem{majhi2019approximating}
Sushovan Majhi, Jeffrey Vitter, and Carola Wenk.
\newblock Approximating {G}romov-{H}ausdorff distance in {E}uclidean space.
\newblock {\em arXiv:1912.13008}, 2019.

\bibitem{manay2006integral}
Siddharth Manay, Daniel Cremers, Byung-Woo Hong, Anthony Yezzi, and Stefano
  Soatto.
\newblock Integral invariants for shape matching.
\newblock {\em Trans. PAMI}, 28:1602--1618, 2006.

\bibitem{memoli2011gromov}
Facundo M{\'e}moli.
\newblock Gromov--{W}asserstein distances and the metric approach to object
  matching.
\newblock {\em Foundations of Computational Mathematics}, 11(4):417--487, 2011.

\bibitem{memoli2022distance}
Facundo M{\'e}moli and Tom Needham.
\newblock Distance distributions and inverse problems for metric measure
  spaces.
\newblock {\em Studies in Applied Mathematics}, 149(4):943--1001, 2022.

\bibitem{memoli2021gromov}
Facundo M{\'e}moli, Zane Smith, and Zhengchao Wan.
\newblock The {G}romov-{H}ausdorff distance between ultrametric spaces: its
  structure and computation.
\newblock {\em arXiv:2110.03136}, 2021.

\bibitem{mosca2020voronoi}
Marco~M Mosca and Vitaliy Kurlin.
\newblock Voronoi-based similarity distances between arbitrary crystal
  lattices.
\newblock {\em Crystal Research and Technology}, 55(5):1900197, 2020.

\bibitem{nigam2022equivariant}
Jigyasa Nigam, Michael~J Willatt, and Michele Ceriotti.
\newblock Equivariant representations for molecular hamiltonians and n-center
  atomic-scale properties.
\newblock {\em Journal of Chemical Physics}, 156(1):014115, 2022.

\bibitem{pomerleau2015review}
Fran{\c{c}}ois Pomerleau, Francis Colas, Roland Siegwart, et~al.
\newblock A review of point cloud registration algorithms for mobile robotics.
\newblock {\em Found. Trends{\textregistered} in Robotics}, 4:1--104, 2015.

\bibitem{pottmann2009integral}
Helmut Pottmann, Johannes Wallner, Qi-Xing Huang, and Yong-Liang Yang.
\newblock Integral invariants for robust geometry processing.
\newblock {\em Comp. Aided Geom. Design}, 26:37--60, 2009.

\bibitem{pozdnyakov2020incompleteness}
Sergey~N Pozdnyakov, Michael~J Willatt, Albert~P Bart{\'o}k, Christoph Ortner,
  G{\'a}bor Cs{\'a}nyi, and Michele Ceriotti.
\newblock Incompleteness of atomic structure representations.
\newblock {\em Phys. Rev. Lett.}, 125:166001, 2020.

\bibitem{rass2022metricizing}
Stefan Rass, Sandra K{\"o}nig, Shahzad Ahmad, and Maksim Goman.
\newblock Metricizing the {E}uclidean space towards desired distance relations
  in point clouds.
\newblock {\em arXiv:2211.03674}, 2022.

\bibitem{rister2017volumetric}
Blaine Rister, Mark~A Horowitz, and Daniel~L Rubin.
\newblock Volumetric image registration from invariant keypoints.
\newblock {\em Transactions on Image Processing}, 26(10):4900--4910, 2017.

\bibitem{ropers2022fast}
Jakob Ropers, Marco~M Mosca, Olga~D Anosova, Vitaliy~A Kurlin, and Andrew~I
  Cooper.
\newblock Fast predictions of lattice energies by continuous isometry
  invariants of crystal structures.
\newblock In {\em International Conference on Data Analytics and Management in
  Data Intensive Domains}, pages 178--192, 2022.

\bibitem{rubner2000earth}
Y. Rubner, C. Tomasi, and L. Guibas.
\newblock The earth mover's distance as a metric for image retrieval.
\newblock {\em Intern. Journal of Computer Vision}, 40(2):99--121, 2000.

\bibitem{schmidt2012learning}
Uwe Schmidt and Stefan Roth.
\newblock Learning rotation-aware features.
\newblock In {\em CVPR}, pages 2050--2057, 2012.

\bibitem{schmiedl2017computational}
Felix Schmiedl.
\newblock Computational aspects of the {G}romov--{H}ausdorff distance and its
  application in non-rigid shape matching.
\newblock {\em Discrete Comp. Geometry}, 57:854--880, 2017.

\bibitem{schoenberg1935remarks}
Isaac Schoenberg.
\newblock Remarks to {M}aurice {F}rechet's article ``{S}ur la definition
  axiomatique d'une classe d'espace distances vectoriellement applicable sur
  l'espace de {H}ilbert.
\newblock {\em Annals of Mathematics}, pages 724--732, 1935.

\bibitem{shi2021robin}
Jingnan Shi, Heng Yang, and Luca Carlone.
\newblock Robin: a graph-theoretic approach to reject outliers in robust
  estimation using invariants.
\newblock In {\em ICRA}, pages 13820--13827, 2021.

\bibitem{simeonov2022neural}
Anthony Simeonov, Yilun Du, Andrea Tagliasacchi, Joshua~B Tenenbaum, Alberto
  Rodriguez, Pulkit Agrawal, and Vincent Sitzmann.
\newblock Neural descriptor fields: {S}{E}(3)-equivariant object
  representations for manipulation.
\newblock In {\em ICRA}, pages 6394--6400, 2022.

\bibitem{sippl1986cayley}
Manfred Sippl and Harold Scheraga.
\newblock Cayley-{M}enger coordinates.
\newblock {\em PNAS}, 83:2283--2287, 1986.

\bibitem{smith2022families}
Philip Smith and Vitaliy Kurlin.
\newblock Families of point sets with identical 1d persistence,.
\newblock {\em arxiv:2202.00577}, 2022.

\bibitem{smith2022practical}
Phil Smith and Vitaliy Kurlin.
\newblock A practical algorithm for degree-k voronoi domains of
  three-dimensional periodic point sets.
\newblock In {\em Lecture Notes in Computer Science (Proceedings of ISVC)},
  volume 13599, pages 377--391, 2022.

\bibitem{spezialetti2019learning}
Riccardo Spezialetti, Samuele Salti, and Luigi~Di Stefano.
\newblock Learning an effective equivariant 3d descriptor without supervision.
\newblock In {\em ICCV}, pages 6401--6410, 2019.

\bibitem{sun2009concise}
Jian Sun, Maks Ovsjanikov, and Leonidas Guibas.
\newblock A concise and provably informative multi-scale signature based on
  heat diffusion.
\newblock {\em Comp. Graph. Forum}, 28:1383--1392, 2009.

\bibitem{toews2013efficient}
Matthew Toews and William~M Wells~III.
\newblock Efficient and robust model-to-image alignment using 3d
  scale-invariant features.
\newblock {\em Medical image analysis}, 17(3):271--282, 2013.

\bibitem{vriza2022molecular}
Aikaterini Vriza, Ioana Sovago, Daniel Widdowson, Peter Wood, Vitaliy Kurlin,
  and Matthew Dyer.
\newblock Molecular set transformer: Attending to the co-crystals in the
  cambridge structural database.
\newblock {\em Digital Discovery}, 1:834--850, 2022.

\bibitem{wang2019deep}
Yue Wang and Justin~M Solomon.
\newblock Deep closest point: Learning representations for point cloud
  registration.
\newblock In {\em Proceedings of CVPR}, pages 3523--3532, 2019.

\bibitem{weisstein2003triangle}
E Weisstein.
\newblock Triangle.
\newblock {\em https://mathworld. wolfram. com}.

\bibitem{wenk2003shape}
Carola Wenk.
\newblock Shape matching in higher dimensions.
\newblock {\em PhD thesis, FU Berlin}, 2003.

\bibitem{widdowson2021pointwise}
D Widdowson and V Kurlin.
\newblock Pointwise distance distributions of periodic point sets.
\newblock {\em arxiv:2108.04798}, 2021.

\bibitem{widdowson2022resolving}
Daniel Widdowson and Vitaliy Kurlin.
\newblock Resolving the data ambiguity for periodic crystals.
\newblock {\em Advances in Neural Information Processing Systems (NeurIPS)},
  35, 2022.

\bibitem{widdowson2022average}
Daniel Widdowson, Marco~M Mosca, Angeles Pulido, Andrew~I Cooper, and Vitaliy
  Kurlin.
\newblock Average minimum distances of periodic point sets - foundational
  invariants for mapping all periodic crystals.
\newblock {\em MATCH Comm. in Math. and in Computer Chemistry}, 87:529--559,
  2022.

\bibitem{zava2023gromov}
N Zava.
\newblock The {G}romov-{H}ausdorff space isn't coarsely embeddable into any
  {H}ilbert space.
\newblock {\em arXiv:2303.04730}, 2023.

\bibitem{zhu2022analogy}
Q Zhu, J Johal, D Widdowson, Z Pang, B Li, C Kane, V Kurlin, G Day, M Little,
  and A Cooper.
\newblock Analogy powered by prediction and structural invariants:
  Computationally-led discovery of a mesoporous hydrogen-bonded organic cage
  crystal.
\newblock {\em J Amer. Chem. Soc.}, 144:9893–9901, 2022.

\bibitem{zhu2022point}
Wen Zhu, Lingchao Chen, Beiping Hou, Weihan Li, Tianliang Chen, and Shixiong
  Liang.
\newblock Point cloud registration of arrester based on scale-invariant points
  feature histogram.
\newblock {\em Scientific Reports}, 12(1):1--13, 2022.

\end{thebibliography}
}

\end{document}